\PassOptionsToPackage{svgnames,table,table,dvipsnames}{xcolor}

\documentclass[journal,hideappendix]{vgtc}        

\onlineid{0}

\vgtccategory{Research}

\vgtcpapertype{evaluation}

\title{Standardness Clouds Meaning: A Position Regarding the Informed Usage of Standard Datasets}

\shortauthortitle{Cech \MakeLowercase{\textit{et al.}}: A Standardness Clouds Meaning: A Position Regarding the Informed Usage of Standard Datasets}

\addtocounter{footnote}{1}
\author{%
  \authororcid{Tim Cech}{0000-0001-8688-2419},
  \authororcid{Ole Wegen}{0000-0002-6571-5897},
  \authororcid{Daniel Atzberger}{0000-0002-5409-7843},
  \authororcid{Rico Richter}{0000-0001-5523-3694},
  \authororcid{Willy Scheibel}{0000-0002-7885-9857}, and
  \authororcid{Jürgen Döllner}{0000-0002-8981-8583}
}

\authorfooter{
\item
    Tim Cech, Ole Wegen, Rico Richter, and Jürgen Döllner are with University of Potsdam, Digital Engineering Faculty. E-mails: \{tcech,wegen,rico.richter.1,doellner\}@uni-potsdam.de
\item
    Daniel Atzberger, and Willy Scheibel are with Hasso Plattner Institute, Digital Engineering Faculty, University of Potsdam. E-mails: \{daniel.atzberger,willy.scheibel\}@hpi.uni-potsdam.de
}

\abstract{Standard datasets are frequently used to train and evaluate Machine Learning models.
However, the assumed \emph{standardness} of these datasets leads to a lack of in-depth discussion on how their labels match the derived categories for the respective use case, which we demonstrate by reviewing recent literature that employs standard datasets.
We find that the standardness of the datasets seems to cloud their actual coherency and applicability, thus impeding the trust in Machine Learning models trained on these datasets.
Therefore, we argue against the uncritical use of standard datasets and advocate for their critical examination instead. For this, we suggest to use \textit{Grounded Theory} in combination with \textit{Hypotheses Testing through Visualization} as methods to evaluate the match between use case, derived categories, and labels.
We exemplify this approach by applying it to the 20 Newsgroups dataset and the MNIST dataset, both considered standard datasets in their respective domain. The results show that the labels of the 20 Newsgroups dataset are imprecise, which implies that neither a Machine Learning model can learn a meaningful abstraction of derived categories nor one can draw conclusions from achieving high accuracy on this dataset.
For the MNIST dataset, we demonstrate that the labels can be confirmed to be defined well.
We conclude that also for datasets that are considered to be \enquote{standard}, quality and suitability have to be assessed in order to learn meaningful abstractions and, thus, improve trust in Machine Learning models.}

\renewcommand{\manuscriptnotetxt}{}

\nocopyrightspace

\graphicspath{{figs/}{figures/}{pictures/}{images/}{./}} 

\usepackage{mathptmx}                  

\DeclareMathAlphabet{\mathcal}{OMS}{cmsy}{m}{n}
\usepackage{numprint}
\npdecimalsign{.} 
\usepackage{booktabs}
\usepackage{tabularx}
\usepackage{csquotes}
\usepackage{amsfonts}
\usepackage{amssymb}
\usepackage{extarrows}
\usepackage[export]{adjustbox}
\usepackage{tikz}
\usetikzlibrary{math,calc,positioning,fit,arrows.meta}
\usepackage{makecell}
\usepackage{fontawesome}
\usepackage{calc}

\usepackage{booktabs}
\usepackage{multirow}   
\usepackage{diagbox}
\usepackage{collcell}
\usepackage{pgfmath}
\usepackage{pgfplots}
\usepackage{pgfplotstable}

\usepackage{listings}

\lstdefinestyle{newsgroupstyle}{
    belowskip=-1.2\baselineskip,
    numbers=left,
    numberstyle=\tiny,
    numbersep=6pt,
    frame=single,
    xleftmargin=12pt,
    basicstyle=\ttfamily\tiny,
    breaklines=true,
    postbreak=\mbox{$\hookrightarrow$\space},
    tabsize=2,
    breakindent=2ex,
}

\newcommand{\citeauthor}[1]{
    \textcolor{red}{ca~\cite{#1}}
}

\pgfplotsset{compat=1.16}

\usetikzlibrary{pgfplots.statistics, pgfplots.colorbrewer}
\usetikzlibrary{matrix}

\colorlet{revision-added}{BrBG-M}
\colorlet{revision-removed}{PiYG-B}
\colorlet{revision-highlight}{Dark2-C}

\usepackage[normalem]{ulem}

\usetikzlibrary{external}
\tikzset{external/system call={pdflatex \tikzexternalcheckshellescape -halt-on-error -interaction=batchmode -jobname "\image" "\texsource"}}
\tikzsetexternalprefix{figures-cached/}
\tikzset{external/only named=true}
\usepackage{xcolor}
\usepackage{pifont}

\usepackage{ifthen}

\newcommand{\CalcCellColor}[1]{%
    \ifthenelse{\equal{\detokenize{#1}}{\detokenize{x}}}
    {}
    {
    \pgfplotsset{colormap/Blues-4}
    \pgfmathfloatparsenumber{#1}%
    \let\value=\pgfmathresult
    \pgfplotscolormapaccess[0:1]{\value}{Blues-4}%
    \xdef\temp{%
        \noexpand\cellcolor[rgb]{\pgfmathresult}%
    }%
    \temp%
    }%
    {#1}
}%

\newcommand{\CalcCellColorInvert}[1]{%
    \ifthenelse{\equal{\detokenize{#1}}{\detokenize{x}}}
    {}
    {
    \pgfplotsset{colormap/Blues-4}
    \pgfmathfloatparsenumber{#1}%
    \let\value=\pgfmathresult
    \pgfplotscolormapaccess[1:0]{\value}{Blues-4}%
    \xdef\temp{%
        \noexpand\cellcolor[rgb]{\pgfmathresult}%
    }%
    \temp%
    }%
    {#1}
}%

\newcolumntype{R}{>{\collectcell{\CalcCellColor}}c<{\endcollectcell}}
\newcolumntype{S}{>{\collectcell{\CalcCellColorInvert}}c<{\endcollectcell}}

\pgfplotstableset{
    /color cells/min/.initial=0,
    /color cells/max/.initial=1000,
    /color cells/textcolor/.initial=,
    color cells/.code={%
        \pgfqkeys{/color cells}{#1}%
        \pgfkeysalso{%
            postproc cell content/.code={%
                \begingroup
                \pgfkeysgetvalue{/pgfplots/table/@preprocessed
                  cell content}\value
                \ifx\value\empty
                    \endgroup
                \else
                \pgfmathfloatparsenumber{\value}%
                \pgfmathfloattofixed{\pgfmathresult}%
                \let\value=\pgfmathresult
                \pgfplotscolormapaccess
                    [\pgfkeysvalueof{/color cells/min}:\pgfkeysvalueof{/color
                      cells/max}]
                    {\value}
                    {\pgfkeysvalueof{/pgfplots/colormap name}}%
                \pgfkeysgetvalue{/pgfplots/table/@cell content}\typesetvalue
                \pgfkeysgetvalue{/color cells/textcolor}\textcolorvalue
                \toks0=\expandafter{\typesetvalue}%
                \xdef\temp{%
                    \noexpand\pgfkeysalso{%
                        @cell content={%
                            \noexpand\cellcolor[rgb]{\pgfmathresult}%
                            \noexpand\definecolor{mapped
                              color}{rgb}{\pgfmathresult}%
                            \ifx\textcolorvalue\empty
                            \else
                                \noexpand\color{\textcolorvalue}%
                            \fi
                            \the\toks0 %
                        }%
                    }%
                }%
                \endgroup
                \temp
                \fi
            }%
        }%
    }
}

\pgfplotstableset{
    colorbarstyle/.style={
        numeric type,
        fixed,
        fixed zerofill,
        precision=1,
        skip 0.=false,
        color cells={min=0,max=1},
        column type=r,
        column type/.add={>{\tiny}}{},
        /pgfplots/colormap name=RdYlBu-CM,
    }
}

\begin{document}


\firstsection{Introduction}
\label{sec:introduction}

\maketitle

\textit{Supervised Machine Learning} algorithms aim to learn a function $f: \mathcal{X} \to \mathcal{Y}$ from a given training dataset $\{(x_1, y_1), \dots, (x_n,y_n)\}$, where $\mathcal{X}$ denotes the feature space of the data point and $Y$ the space of target values~\cite{mohri2018foundations}.
For example, in case of a regression model, the target values are real numbers, i.e., $\mathcal{Y} = \mathbb{R}$, and the feature space would be given by real $n$-tuples, i.e., $\mathcal{X} = \mathbb{R}^n$.
The quality of a model is described via a cost function, that quantifies how well the parameters fit the training data.
By optimizing the cost function, e.g., by gradient descent, the parameters of the function $f$ are updated to fit the data better.
After fitting the model on the training data, the performance can be further evaluated on a previously unseen test data set.
If the model performs well on the test data, it is usually deployed in a real-world application.

\begin{figure}[t]
    \centering%
    \includegraphics[width=1.0\linewidth]{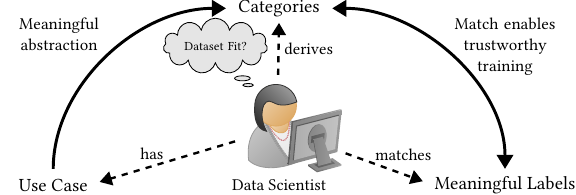}
    \caption{We argue that a data scientist must actively ensure the match between the use case, labels, and categories (dashed lines). The use case provides the context for deriving categories which should match the labels (solid lines).}
    \label{fig:model}%
\end{figure}

Although the effectiveness of the model in application depends largely on the content and quality of the training data, studies often neglect a detailed initial examination of this data.
This harbors the risk that the model learns unwanted patterns that do not correspond to the expectations. This is particularly relevant for so-called \enquote{standard} datasets, which are often used as a baseline.
For text classification, one such standard dataset is the \textit{20 Newsgroups} dataset~\cite{non_discussion_1,non-discussion_3}, while for computer vision, an example is the \textit{MNIST (Modified National Institute of Standards and Technology)} dataset~\cite{chen2018assessing, garg2019validation}.
We perceive that standard datasets are often used uncritically based on their \emph{standardness}, which clouds two possible breaking points:
\begin{itemize}
\item Mismatch between labels and concepts: We refer to the elements of $Y$ as labels, i.e., pre-defined classes, and the concepts as the underlying ideas conveyed by the name of the labels.
\item Unsuitability for the use case: Even if labels and corresponding concepts match, a dataset can be unsuited for a given application, e.g., when using a text corpus containing excerpts from novels for training a model that should later classify news articles.
\end{itemize}
In our view, these two possible breaking points are often not discussed (as we will demonstrate), relying on and appealing to the standardness of the used datasets, which, however, is an unclear characterization, especially since there is no qualitative process of how a dataset becomes \enquote{standard}.
Often, this happens due to historical developments and the subsequent frequent use of a dataset, which provides no guarantee for the quality of labels and the applicability to specific use cases.
Therefore, also standard datasets have to be closely examined, as
their labels only convey \textit{meaning} if they adhere to the (humanly understanded) match between use case, derived categories, and class labels (see \autoref{fig:model}).
If this is not the case, the result is uninterpretable and untrustworthy, as it is unclear what the ML model learned~\cite{entitlementsML}.
Therefore, trustworthy reasoning about the quality of an ML model can not rely on quantifiable metrics (e.g., accuracy) only, but also requires the critical assessment of the used (standard) datasets.


In this paper, we suggest to adopt interdisciplinary methods, namely \textit{Grounded Theory} \cite{glaser2017discovery} in combination with \textit{Hypotheses Testing through Visualization} to inspect a given training dataset as quali-quantitative methods \cite{quali_quantitative_methods} for dataset interrogation.
We exemplify this approach by applying it to the 20 Newsgroups dataset as a negative case (i.e., labels do not match categories and/or the dataset is not suitable for the targeted use case) and the MNIST dataset as a positive case (i.e., labels match categories and the dataset is suitable for the targeted use case).
In summary, as our main contribution, we question the term of \emph{standardness} in relation to datasets, which often clouds the need to explicitly match use cases, categories and class labels in order to achieve explainability and trustworthiness.

The remaining part of this work is structured as follows:
In \autoref{sec:DatasetAssessment}, we will motivate the importance of dataset assessment in general and specifically for standard datasets. For this, we discuss dataset quality and suitability in the context of ML.
In \autoref{sec:standardness}, we describe the characteristics of the 20 Newsgroups and MNIST datasets and demonstrate their uncritical use in literature.
In \autoref{sec:method}, we suggest a quali-quantiative method for manual data assessment, for which we exemplify the application on the two datasets in \autoref{sec:case study}.
We discuss and summarize our findings in \autoref{sec:Discussion} and \autoref{sec:Conclusions}.

\section{The Importance of Dataset Assessment}
\label{sec:DatasetAssessment}

Datasets are the foundation of ML methods and it is now well understood that ML model performance is \enquote{upper bounded by the quality of the data}~\cite[p.1]{jain2020dataQuality}.
Furthermore, algorithm audits revealed that ML models have the potential to be racist and sexist~\cite{sexist_algorithms1}, can threaten users~\cite{threat_of_invisibility}, and can produce unfair and biased output~\cite{survey_fairness}.
The reason for these negative effects is  often connected to the training data used and could be avoided by assessing their quality and suitability beforehand.

\subsection{Adressing Dataset Quality}
The issue of data quality for ML has been discussed more thoroughly in recent years~\cite{priestley2023dataQuality} and recommendations for building high quality datasets for use in ML have been proposed, e.g., the dataset definition standard of Cappi et al.~\cite{cappi2021datasetDefintionStandard} and Picard et al.~\cite{picard2020datasetQuality} that demands, among others, representativeness, traceability, reliability, label consistency and accuracy, as well as minimized bias, or the overview of data measurement methods for ML provided by Mitchell et al.~\cite{mitchell2022measuringData}.
In the context of the dataset nutrition label project, a label was proposed for the overview of dataset \enquote{ingredients}, comprising qualitative and quantitative information~\cite{holland2020datasetNutritionLabel}.
Giner-Miguelez et al. perceive that after a lot of model-centric research had been conducted, we now experience a \enquote{data-centric cultural shift, where data issues are given the attention they deserve and more standard practices for gathering and describing datasets are discussed and established}~\cite[p.1]{miguelez2023dlsForMLDatasets}.

However, the question arises of how to treat existing datasets that did not use the proposed practices and documentation methods.
For this task, methods for data quality interventions have been proposed that can be used to improve existing datasets, e.g., using interactive data cleaning systems, driven by integrity constraints, schema matching, or outlier detection~\cite{sambasivan2021dataWork}.
However, a lot of earlier research focused on quantitative, model-centric aspects. 
In this context, the field of \textit{Explainable AI (XAI)} emerged~\cite{explainability_cs}, and while acknowleding the issue of biases~\cite{survey_fairness}, attempted to provide automatic explanations for opaque decision-making ML models on the basis of quantitative analysis.
Thus, XAI usually does not engage in qualitative, manual analysis of the data~\cite{quali_quantitative_methods, lazer2014parable, entitlementsML}.
However, in spite of this profoundly model-centric approach, some XAI research also attempts to understand noise in labels on a qualitative level~\cite{pmlr-v97-chen19g}.

Another data-centric approach is confidence learning as a subfield of ML, which is concerned with potential errors in labels and how to mitigate their impact.
The field is constituted by the observation that \enquote{Advances in learning with noisy labels [...] usually introduce a new model or loss function.
Often, this model-centric approach band-aids the real question: which data is mislabeled?}~\cite[p.1]{northcutt2021confident}.
Typically, the noise in labels is modeled as a random variable to account for it during training~\cite{pmlr-v97-chen19g,natarjan_cl,northcutt2021confident}.
However, we want to emphasize that qualitative noise may be intractable by a simple mathematical quantification, e.g., a random variable.
We argue that manual investigation is required first to determine whether the alleged noise in labels is, in reality, a systemic, frequent gap between the labels and the categories (meaning) of the data. 
Only after this step, the actual noise in labels can be modeled correctly.

A qualitative view on the matter of datasets is helpful here. 
Neff et al. state that datasets are \enquote{the media through which conversation, negotiation, and action can occur}~\cite[p.2]{neff_critique_and_contribute}.
Here, datasets are perceived as basis for automated decision-making and, thus, in a way exercise power over people or shape their perception of reality~\cite{power_1,power_2}.
A particular focus is placed on the fact that it is impossible to treat ML models (and, therefore, implicitly their training datasets) objectively and separate them from their societal context as the public tends to mystify them~\cite{ai_as_myth1} and programmers can usually not restrict their ML models to the computational context~\cite{seaver2017algorithms} or avoid making ethical decisions~\cite{mittelstadt2016ethics}. A purely quantitative approach to dataset assessment is, therefore, unsuitable and qualitative dataset interrogation is required to get a comprehensive view of a dataset and its context.

\subsection{Addressing Dataset Suitability}
Besides the issue of data quality, another important aspect is the suitability of a dataset for a specific use case. 
In this context, information on the provenance of the data is crucial.
In the domain of database management, it has been clear for years that \enquote{it is of utmost importance to understand the provenance of data} in order to assess its \enquote{quality and trustworthiness}~\cite[p.381]{cheney2009provenance}.
However, until recently, data provenance had rarely been discussed in the ML community, despite the critical and foundational role of datasets for ML. 
Therefore, Gebru et al. propose that every dataset should be documented regarding motivation, composition, collection process, and recommended uses~\cite{gebru2021datasheetsForDatasets} in order to facilitate, among others, the selection of appropriate datasets for a given task.
However, it has to be taken into account that the suitability of a dataset for a given task may change over time, as supervised models face the fundamental problem that they become outdated once the training data does not conform with the observed patterns any longer. 
This is, again, due to the fact that the models rely mainly on inductive reasoning~\cite{review_epistomological_foundations} while encoding the (historical) societal context in which the training data emerged.

\subsection{Standardness Does not Equal Quality}
While more data-centric research on ML has been conducted in recent years, we perceive that for so-called standard datasets the issues of quality and suitability are rarely taken into account, which we will demonstrate in case studies below. 
The recently proposed recommendations and processes for obtaining high-quality, well-documented datasets do not apply here, as standard datasets already exist, often for many years. 
The methods for handling existing datasets are rarely used, as if the standardness implies quality and suitability, which is not necessarily true.
The etymology of the word \enquote{standard} provides a good analogy here. 
The word is derived from the old french word \enquote{estandart} which denotes a military banner or flag, positioned by military authorities, and used as rallying point (\enquote{that to which one turns} according to the Century Dictionary Volume VII, p. 5900).
In the same sense, the standardness of a dataset usually means that many researchers turn to this dataset, i.e., use it for training and evaluation of ML models. 
Ideally, this standard dataset would serve as a reliable and suitable rallying point, to speak in the analogy. 
However, in contrast to the military banner, this is not guaranteed, as there is no established process for how a dataset becomes \enquote{standard}.
Often, this happens due to historical developments (e.g., first dataset in an area or influential paper that uses this dataset) and the subsequent frequent use of a dataset.
Therefore, we claim that the standardness of a dataset does not guarantee a certain data quality with respect to correct labels, noise, or diversity.
Consequently, we argue for the critical assessment and use of standard datasets, not relying on their standardness alone. 
Due to the unclear definition and characteristics of standard datasets, every researcher should examine the used datasets carefully and with respect to the targeted use case, combining quantitative and qualitative methods.
While this may seem obvious, we perceived that researchers often use standard datasets without a careful examination of their quality and an in-depth discussion about why the datasets are appropriate for the intended use case.

\section{Researchers' Questionable Reliance on Standardness}
\label{sec:standardness}

To underpin and exemplify our claims, we now provide a descriptive analysis of two datasets, namely the 20 Newsgroups dataset and the MNIST dataset, to illustrate possible issues with these datasets in general, despite their standardness. 
As shown below, both datasets are older.
Naturally, competitors to both datasets arose in the meantime, but for them, it is too early to claim as clearly that any of them has already become a new standard, especially considering both datasets are still used in recent papers.
For each dataset, we review 15 research papers that make use of it and evaluate how well the use of the dataset is motivated and how well the dataset itself is interrogated.
We included a paper if either a model is evaluated on the dataset or the dataset is heavily used otherwise.
We chose the top papers according to Google Scholar with the cut-off date at 01-01-2015 to review reasonably recent research articles.
We want to clarify that gathering sources by Google Scholar is not deterministic but often good enough to find pertinent references~\cite{giustini2013google}.
Therefore, our review is not replicable and can not be considered a systematic survey.
Nevertheless, we settled for this method since, in our experience, computer science researchers still use Google Scholar to find relevant references and, as far as scientific literature is concerned, researchers are likely introduced to the datasets through papers they discovered using this search engine.

We assess each paper with respect to five categories, namely 
(1) whether the author(s) of the research paper gave a traceable reference to the dataset, 
(2) whether they explicitly chose a specific version of the dataset, 
(3) whether they communicated the content of the dataset at all, 
(4) whether they gave an example from the dataset, and 
(5) whether they performed an in-depth analysis of the contents for
their use case (suitability).
These categories characterize the level of detail of the assessment of the dataset.
The results of the review are shown in \autoref{tab:review} and \autoref{tab:review_mnist}.

\subsection{20 Newsgroups Dataset}
The 20 Newsgroups dataset contains texts sent via e-mail lists in the late 1990s curated by Ken Lang \cite{joachims1996probabilistic}.
We obtained the 20 Newsgroups dataset from \href{http://qwone.com/~jason/20Newsgroups/}{Jason Rennie's Home Page}\footnote{\href{http://qwone.com/~jason/20Newsgroups/}{http://qwone.com/~jason/20Newsgroups/}} since it often appears to be given as the source \cite{scikit_source,scikit_source2}.
Additionally, according to their respective documentation, the popular Python ML library \href{https://scikit-learn.org/stable/datasets/real_world.html#the-20-newsgroups-text-dataset}{Scikit-Learn}\footnote{\href{https://scikit-learn.org/stable/datasets/real_world.html\#the-20-newsgroups-text-dataset}{https://scikit-learn.org/stable/datasets/real\_world.html\#the-20-newsgroups-text-dataset}} and the dataset-sharing site \href{https://www.kaggle.com/datasets/crawford/20-newsgroups}{Kaggle}\footnote{\href{https://www.kaggle.com/datasets/crawford/20-newsgroups}{https://www.kaggle.com/datasets/crawford/20-newsgroups}} also state this source.
According to Jason Rennie's website, three different versions of the dataset exist.
The original dataset contains \numprint{19997} documents, approximately evenly distributed over 20 categories.
A second version with fewer headers and removed duplicates contains \numprint{18846} documents.
A third version with even less duplicates but maintained \textit{From} and \textit{Subject} header lines contains \numprint{18828} documents.
The website speculates that the dataset was gathered for the work of Lang \cite{Lang95} though it was finally not included in the paper.
This proposition is strengthened by other sources from the 90s, for example by Joachims et al. \cite{joachims1996probabilistic}.

The categories are constituted by the name of the respective mailing list.
Each category contains a selection of e-mails from several threads that appear to be loosely connected.
According to the header field \textit{From}, the dataset contains posts from \numprint{8644} unique e-mail addresses.
Please note that we assumed that an author is uniquely identified by the content of the \emph{From} line.
We ignored the case where one author uses several aliases or e-mail addresses.
Naturally, not all authors contributed equally to the dataset.
Only \numprint{3080}, \numprint{1659}, and \numprint{293} authors contributed at least two, three, and ten posts, respectively.
This corresponds to a share of \numprint[\%]{72}, \numprint[\%]{57}, and \numprint[\%]{27} of the number of all posts from \numprint[\%]{36}, \numprint[\%]{19}, and \numprint[\%]{3} of all authors, respectively.
At the line level, the difference between the number of total lines (\numprint{762321}) and lines of authors with at least two (\numprint{601304} lines corresponding to a share of \numprint[\%]{78}), three (\numprint{505800} lines corresponding to a share of \numprint[\%]{66}), and ten (\numprint{258885} lines corresponding to a share of \numprint[\%]{34}) posts becomes even more apparent.
Therefore, the language of only three percent of authors represents a disproportional share of around one-third of the text corpus.
Furthermore, we also want to emphasize that much of the content comprises quotes from other e-mails, indicated by a starting right arrow symbol.
Overall, over \numprint[\%]{16.5} of all lines (\numprint{126202} lines) are quotes and not original content from the respective author.
The high ratio of quotes results in texts that significantly differ from spoken language~\cite{on_quotes}.

Sometimes, e.g., by Albishre et al.~\cite{basic_discussion1} or Zhang and Yamana \cite{zhang_and_yamana}, the already mentioned paper by Lang \cite{Lang95} is given as the source. 
Curiously, Kou et al.~\cite{KOU2020105836} state that Joachims et al.~\cite{joachims1996probabilistic} collected the 20 Newsgroups dataset, which is doubtful since Joachims et al. themselves gives Lang as the original collector~\cite{joachims1996probabilistic}.
Saigal et al.~\cite{saigal_et_al} link to \href{https://www.ics.uci.edu/}{another website}\footnote{\href{https://www.ics.uci.edu/}{https://www.ics.uci.edu/}} and additionally state without reference that the dataset was collected by Lang.
Hsu \cite{hsu} references the Scikit-Learn library as the source, which references the original website.
Often, Jason Rennie's website is identified as the source of the dataset~\cite{almost_no_discussion_1,non-discussion_3,non_discussion_1,scikit_source,scikit_source2}.
Seldom, the different versions of the dataset are mentioned to make transparent which version of the dataset was used (e.g., by Albshire et al.~\cite{basic_discussion1}).
More often, the authors state that the dataset contains approximately \numprint{20000} documents from 20 different classes which holds for all three versions of the dataset (e.g., in Wang et al.~\cite{non_discussion_1}, Pappagari et al~\cite{non-discussion_3}, and Chandra~\cite{almost_no_discussion_1}).
Interestingly, the authors rarely, and then only briefly, discuss what they think the 20 Newsgroups contain.
Usually, the discussion tends to be on the descriptive level.
It contains statements about the context and lists some example label names~\cite{KOU2020105836,saigal_et_al}.
Only in one case, an example document was shown, however, without discussion~\cite{basic_discussion1}. 
Furthermore, no reviewed paper discusses in detail why the usage of the 20 Newsgroups dataset is appropriate for the intended use case.
In summary, in the reviewed literature it remains unclear why the 20 Newsgroups dataset was used and why the authors validate their method on this dataset.

\begin{table}[t]
    \centering
    \pgfplotstableread[col sep=comma]{tables/Paper_20_newsgroups.csv}\newsgroupdata
    \pgfplotstabletypeset[
        columns={
            Paper,
            Source?,
            Explicit Choice?,
            States Content?,
            Example?,
            In-Depth?
        },
        columns/Paper/.style={
            string type,
            column type={X},
            column name={ \textbf{Paper} },
        },
        columns/Source?/.style={
            string type,
            column type={c},
            column name={ \rotatebox{90}{\textbf{Source Given?}} },
        },
        columns/Explicit Choice?/.style={
            string type,
            column type={c},
            column name={ \rotatebox{90}{\textbf{Explicit Choice?}} },
        },
        columns/States Content?/.style={
            string type,
            column type={c},
            column name={ \rotatebox{90}{\textbf{Content Stated?}} },
        },
        columns/Example?/.style={
            string type,
            column type={c},
            column name={ \rotatebox{90}{\textbf{Example Given?}} },
        },
        columns/In-Depth?/.style={
            string type,
            column type={c},
            column name={ \rotatebox{90}{\textbf{Suitability Analyzed?}} },
        },
        font=\footnotesize,
        col sep=&,
        row sep=\\,
        every head row/.style={
            before row=\toprule,
            after row=\midrule
        },
        every last row/.style={after row=\bottomrule},
    	every odd row/.style={before row={\rowcolor[gray]{.90}}},
        begin table={
           \begin{tabularx}{\linewidth}
        },
        end table={
           \end{tabularx}
        },
    ]{\newsgroupdata}
    \caption{An overview of the 15 reviewed ML research papers for the 20 Newsgroups dataset, assessed according to the defined categories. 
    We used the entry \emph{(yes)} if the property was given indirectly.}
    \label{tab:review}
\end{table}

\subsection{MNIST Dataset}

The MNIST dataset contains 28x28 pixel images of hand-written digits from 2000 Census Bureau employees and 500 high school students from the 1990s~\cite{baldominos_good_survey,chhavi_reconstruction}.
Baldominos et al.~\cite{baldominos_good_survey} state that the dataset origins from the NIST special datasets 1 and 3 which were later combined into special dataset 19~\cite{grother1995nist}.
In contrast to the 20 Newsgroups dataset, the history of the MNIST dataset is documented by the study of Chhavi and Bottou~\cite{chhavi_reconstruction}.
According to them, Bottou et al.~\cite[p.3]{mnist_original_original} introduced the dataset in 1994.
However, since this conference paper was added only in 2002 to the \href{https://ieeexplore.ieee.org/document/576879}{IEEE digital library}\footnote{\href{https://ieeexplore.ieee.org/document/576879}{https://ieeexplore.ieee.org/document/576879}}, eight years after it was originally published, the origin of the dataset is often wrongly contributed to LeCun et al.~\cite{lecun1998gradient} of the same working group~\cite{baldominos_good_survey,cohen2017emnist,Serrano-Gotarredona_MNIST}.
Since most papers reference LeCun et al.~\cite{lecun1998gradient} as the dataset's source and \href{http://yann.lecun.com/exdb/mnist/}{LeCun's website}\footnote{\href{http://yann.lecun.com/exdb/mnist/}{http://yann.lecun.com/exdb/mnist/}} claims that it contains the dataset referenced in the paper, we assume that this version of the dataset is the authoritative source.
Notably, neither LeCun's website nor LeCun et al.~\cite{lecun1998gradient} reference Bottou et al.~\cite{mnist_original_original}.
We additionally want to note that according to Chhavi and Bottou~\cite[p.3]{chhavi_reconstruction} \enquote{although [Bottou et al.~\cite{mnist_original_original}] was the first paper mentioning MNIST, the creation of the dataset predates this benchmarking effort by several months} and when the paper was written, the dataset was already fully established.
Furthermore, parts of the dataset appear in other papers that predate the paper from 1994~\cite{denker_digits,le_cun_digits}.
Consequently, Chhavi and Bottou~\cite{chhavi_reconstruction} state that the exact origin of the MNIST dataset remains unclear.

The categories of the dataset are constituted by forms filled out by Census Bureau employees and students.
An example form is shown on the corresponding \href{https://www.nist.gov/srd/nist-special-database-19}{NIST dataset site}\footnote{\href{https://www.nist.gov/srd/nist-special-database-19}{https://www.nist.gov/srd/nist-special-database-19}} or in Grother and Hanaoka~\cite{grother1995nist}.
Since this data was gathered in a highly controlled environment, the labels are much more reliable than the 20 Newsgroups dataset \cite{grother1995nist}.
However, since the original form fields contained number sequences, each of these sequences had to be split to obtain a dataset of single digits.
The algorithm by which the digit sequence was broken down and preprocessed into single digits could not be reverse-engineered by a team member from the original 1994 paper under the consultation of several other team members \cite{chhavi_reconstruction}.
Further, the authors discovered bugs that resulted, for example, in a pixel shift for certain digits that impacted model testing errors when corrected~\cite{chhavi_reconstruction}.

The results from the review in \autoref{tab:review_mnist} are more promising compared to the results for the 20 Newsgroups dataset.
Usually, the authors indirectly reference the single dataset version on LeCun's website~\cite{baldominos_good_survey,cohen2017emnist,castro2019morpho}.
Only once in Chhavi and Bottou \cite{chhavi_reconstruction}, the earlier 1994 paper is mentioned, which, however, did not publish the used dataset version \cite{mnist_original_original}.
Additionally, the actual contents of the dataset are discussed more frequently compared to the 20 Newsgroups dataset and example images are shown~\cite{baldominos_good_survey,kadam2020cnn,chhavi_reconstruction}.
Castro et al. \cite{castro2019morpho} only describe abstractly what the MNIST dataset is and mainly rely on the references LeCun et al. \cite{lecun1998gradient} and Chhavi and Bottou \cite{chhavi_reconstruction}.
With Chhavi and Bottou~\cite{chhavi_reconstruction} and Baldominos et al.~\cite{baldominos_good_survey}, two papers discuss specifics of the dataset.
Nevertheless, papers often fail to discuss in detail why they find the MNIST dataset appropriate for evaluating their model \cite{castro2019morpho,garin2019topological,saadna2019speed}.
If the scope of a paper is limited to the usage of the MNIST dataset specifically, this may be not as important as for a more general use case.
Nevertheless, it would be important to discuss, why this limitation leads to relevant results.

In summary, the MNIST dataset is better documented compared to the 20 Newsgroups dataset, even though some details must be considered lost in history.
Also, the labels are more reliable since they were specifically captured by government employees and high school students in a dedicated form.

\begin{table}
    \centering
    \pgfplotstableread[col sep=comma]{tables/Paper_20_mnist.csv}\mnistdata
    \pgfplotstabletypeset[
        columns={
            Paper,
            Source?,
            Explicit Choice?,
            States Content?,
            Example?,
            In-Depth?
        },
        columns/Paper/.style={
            string type,
            column type={X},
            column name={ \textbf{Paper} },
        },
        columns/Source?/.style={
            string type,
            column type={c},
            column name={ \rotatebox{90}{\textbf{Source Given?}} },
        },
        columns/Explicit Choice?/.style={
            string type,
            column type={c},
            column name={ \rotatebox{90}{\textbf{Explicit Choice?}} },
        },
        columns/States Content?/.style={
            string type,
            column type={c},
            column name={ \rotatebox{90}{\textbf{Content Stated?}} },
        },
        columns/Example?/.style={
            string type,
            column type={c},
            column name={ \rotatebox{90}{\textbf{Example Given?}} },
        },
        columns/In-Depth?/.style={
            string type,
            column type={c},
            column name={ \rotatebox{90}{\textbf{Suitability Analyzed?}} },
        },
        font=\footnotesize,
        col sep=&,
        row sep=\\,
        every head row/.style={
            before row=\toprule,
            after row=\midrule
        },
        every last row/.style={after row=\bottomrule},
    	every odd row/.style={before row={\rowcolor[gray]{.90}}},
        begin table={
            \begin{tabularx}{\linewidth}
        },
        end table={
            \end{tabularx}
        },
    ]{\mnistdata}
    \caption{An overview of the 15 reviewed ML research papers for the MNIST dataset, assessed according to the defined categories. 
    We used the entry \emph{(yes)} if the property was given indirectly.
    }
\label{tab:review_mnist}
\end{table}

\subsection{Summary}
We have exemplified that recent papers frequently omit the discussion of whether a dataset is suitable for evaluating a model.
In the case of the 20 Newsgroups dataset, none of the considered papers critically discussed the use of the dataset.
In the case of the MNIST dataset, Chhavi and Bottou~\cite{chhavi_reconstruction} and Baldominos et al.~\cite{baldominos_good_survey} provide such a critical discussion, but in recent papers this discussion is either not reflected, e.g., in Kadam et al.~\cite{kadam2020cnn} or Cheng and Hayato~\cite{cheng2020analysis}, or only briefly reflected, e.g., in Castro et al.~\cite{castro2019morpho}.
In our opinion, this lack of discussion is often induced by relying on the standardness of a dataset. 
Instead, a dataset should be carefully evaluated and not used uncritically.
In addition to descriptive analysis as demonstrated above, dataset interrogation via manual assessment is required to get a comprehensive view of a dataset.
To clarify what we mean by that and to aid researchers in this manual assessment, we will now suggest a quali-quantiative method~\cite{quali_quantitative_methods} and exemplify its use on the 20 Newsgroups and MNIST datasets.


\section{A Quali-Quantitative Method for Dataset Interrogation}
\label{sec:method}
 
Seth et al. argue that in addition to automated quantitative evaluation, one should use manual content evaluation to capture \emph{latent features} that may not be as easily quantified~\cite{manual_investigation}.
Similarly, Blok and Pedersen~\cite{quali_quantitative_methods} argue that qualitative methods from social science can complement the quantitative methods that are commonly used in computer science.
In this regard, Cech et al.~\cite{cech_quali_quantitative} have given a first impression of how such mixed methods can lead to new knowledge artifacts about ML models.
We want to suggest a quali-quantiative approach to examining datasets, comprising Grounded Theory and Hypothesis Testing through Visualization.

\subsection{Grounded Theory}
Grounded Theory is a qualitative method introduced to sociological research by Glaser and Strauss~\cite{glaser2017discovery} and recently also discussed in computer science research~\cite{diehl2022characterizing}.
Grounded Theory entails, among others, \textit{Close Reading}, \textit{Coding}, and \textit{Theoretical Sampling} of datasets and literature~\cite{engler2011grounded, diehl2022characterizing}.
Close Reading describes the process of looking at data points in detail by engaging in manual evaluation~\cite{diehl2022characterizing}.
Coding describes the process of mapping investigated data points to a predefined set of descriptive categories~\cite{engler2011grounded}.
Theoretical Sampling, in contrast to random sampling, is the process of acquiring more data based on the current form of one's theory~\cite{engler2011grounded}.
This process is often not linear and requires constant readjustments of one's theory so that it becomes compatible with the newly found data.
This process is reiterated until no new knowledge emerges and, as such, the theory reaches so called \textit{Theoretical Saturation}~\cite{diehl2022characterizing}.

\subsection{Hypothesis Testing through Visualization}
As Grounded Theory is a qualitative method by nature, it requires manual investigation per datapoint~\cite{diehl2022characterizing}.
This is not feasible for the majority of data points in case of large datasets.
Therefore, we complement this approach with an additional method that employs a quantitative technique to strengthen (or counter) the findings from applying Grounded Theory.
Statistical hypothesis testing is an established quantitative method in different research areas~\cite{statistical_testing}.
It starts from a so-called null hypothesis with a qualitative proposition and subsequently examines available data to find out, whether the null hypothesis has to be rejected, assuming a certain quantitative distribution.
We want to take up this method as an analogy to propose \textit{Hypothesis Testing through Visualization}, situated in the broader context of \textit{VIS4GT (Visualization for Grounded Theory)} as introduced by Diehl et al.~\cite{diehl2022characterizing}.

If visualization is informed by prior qualitative findings, it can enrich the understanding of and trust in ML models~\cite{quality_helps_visualization,human_informed_dr} and can help in examining whether a model has learned meaningful concepts from the training data.
Therefore, we suggest to use a quantitative visualization technique of the data (\textit{Dimensionality Reduction (DR)} into a two-dimensional scatter plot), to examine whether the data points follow the expectations with respect to the hypothesis.
This mirrors the assumed quantitative distribution from the classical hypothesis testing.
We argue that a purely qualitative hypothesis (obtained via Grounded Theory) is required first to mitigate the risk that the subsequent interrogation of the dataset is distorted by the quantitative visualization technique.
Similar to a null hypothesis, the opposite of the hypothesis found via Grounded Theory should be chosen initially.

However, as recent work by Jeon et al.~\cite{jeon2023classes} points out, care has to be taken when using DR techniques in the context of visual analysis, as clusters in the two-dimensional plane can not easily be assumed to represent meaningful categories.
More likely, categories are spanning an arbitrary structure in the high-dimensional space, making it difficult to find appropriate projections that lead to interpretable results.
While we agree, we investigate the problem at a more fundamental level, asking whether a concrete class label represents a coherent category in the high-dimensional space at all.
For this, we rely on benchmarking studies, e.g., Espadoto et al.~\cite{espadoto2019toward} and Atzberger and Cech et al.~\cite{benchmarking}, to choose appropriate methods for DR that lead to reliable representations.
Atzberger and Cech et al.~\cite{benchmarking} focused on text data while Espadoto et al.~\cite{espadoto2019toward} employed a more general approach.
While this approach may obtain a sub-optimal but still \enquote{good enough} layout, it mitigates the risk of over-optimization~\cite{googd_enough_vis}.



\begin{figure}[t]
    \centering
\begin{lstlisting}[language={},style=newsgroupstyle,linerange={25-50}]
                          An Introduction to Atheism
                       by mathew <mathew@mantis.co.uk>

This article attempts to provide a general introduction to atheism.  Whilst I
have tried to be as neutral as possible regarding contentious issues, you
should always remember that this document represents only one viewpoint.  I
would encourage you to read widely and draw your own conclusions; some
relevant books are listed in a companion article.

To provide a sense of cohesion and progression, I have presented this article
as an imaginary conversation between an atheist and a theist.  All the
questions asked by the imaginary theist are questions which have been cropped
up repeatedly on alt.atheism since the newsgroup was created.  Some other
frequently asked questions are answered in a companion article.

Please note that this article is arguably slanted towards answering questions
posed from a Christian viewpoint.  This is because the FAQ files reflect
questions which have actually been asked, and it is predominantly Christians
who proselytize on alt.atheism.

So when I talk of religion, I am talking primarily about religions such as
Christianity, Judaism and Islam, which involve some sort of superhuman divine
being.  Much of the discussion will apply to other religions, but some of it
may not.

"What is atheism?"
\end{lstlisting}%
    \caption{Document \numprint{51060} of the category alt.atheism. The text contains an article about atheism and therefore has a direct link to the category label.}
    \label{fig:good_atheism_1}
\end{figure}

\begin{figure}[t]
    \centering
\begin{lstlisting}[language={},style=newsgroupstyle,linerange={15-31}]
>Why is it more reasonable than the trend towards obesity and the trend towards
>depression? You can't just pick your two favorite trends, notice a correlation 
>in them, and make a sweeping statement of generality. I mean, you CAN, and 
>people HAVE, but that does not mean that it is a valid or reasonable thesis. 
>At best it's a gross oversimplification of the push-pull factors people 
>experience.  

I agree, I reckon it's television and the increase in fundamentalism.. You
think its the increase in pre-marital sex... others thinks its because
psychologists have taken over the criminal justice system and let violent
criminals con them into letting them out into the streets... others think
it's the increase in designer drugs... others think it's a communist plot.
Basically the social interactions of all the changing factors in our society
are far too complicated for us to control. We just have to hold on to the
panic handles and hope that we are heading for a soft landing. But one
things for sure, depression and the destruction of the nuclear family is not
due solely to sex out of marriage.
\end{lstlisting}%
    \caption{Document \numprint{51194} of the class label alt.atheism. The text is part of a discussion about whether the growing number of people identifying as atheists is correlated to the growing number of depression cases.}
    \label{fig:good_atheism_2}
\end{figure}

\begin{figure}[t]
    \centering
    \frame{\includegraphics[width=1.0\columnwidth]{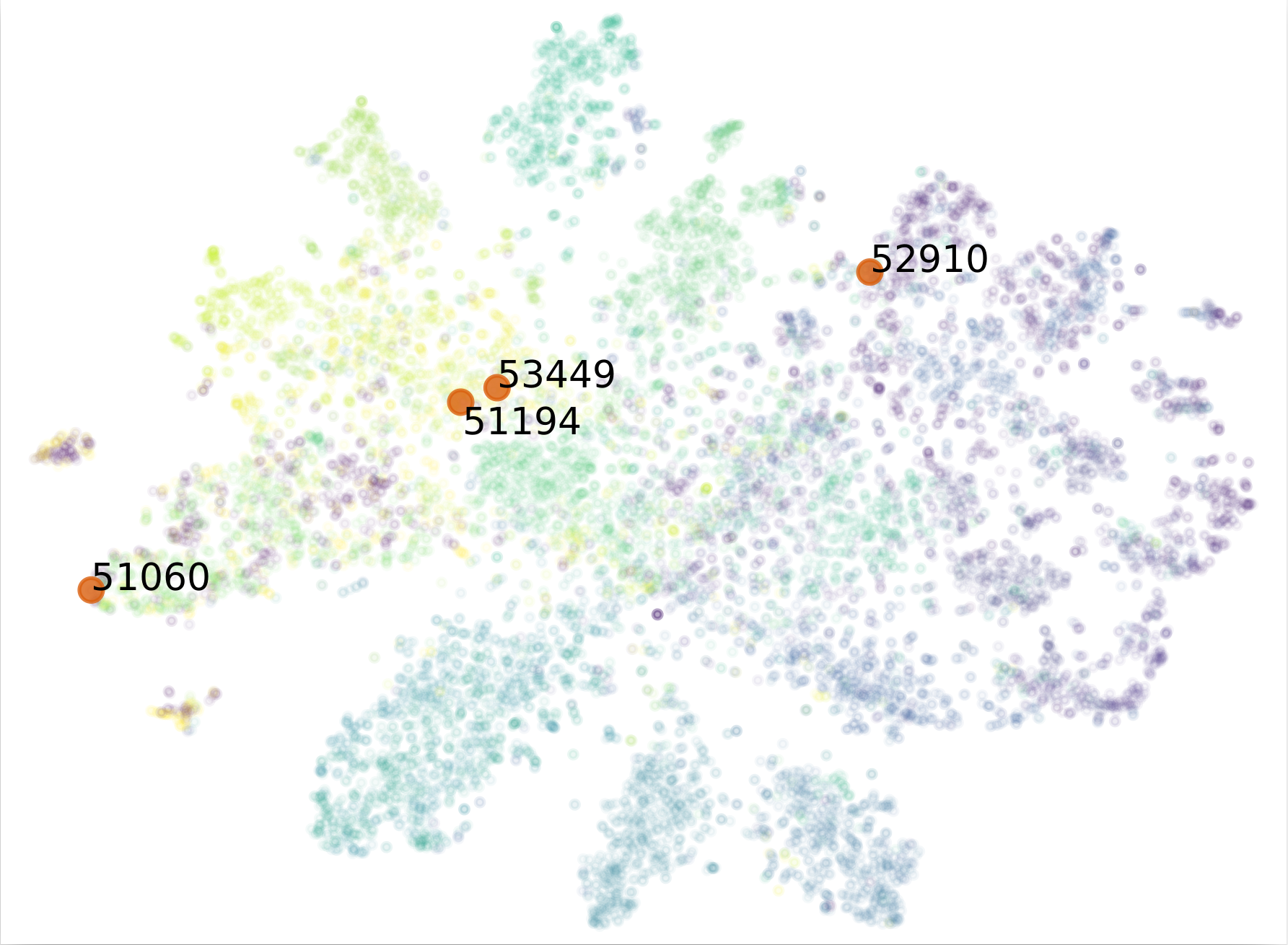}}%
    \caption{The 20 Newsgroups dataset was reduced with t-SNE and LSI as proposed by Atzberger and Cech et al. \cite{benchmarking}. The documents \numprint{51060}, \numprint{51194}, \numprint{52910}, and \numprint{53449} are highlighted. They are wide-spread in the visualization and are also semantically dissimilar.}
    \label{fig:lsi}
\end{figure}

\begin{figure}[t]
    \centering
    \frame{\includegraphics[width=1.0\columnwidth]{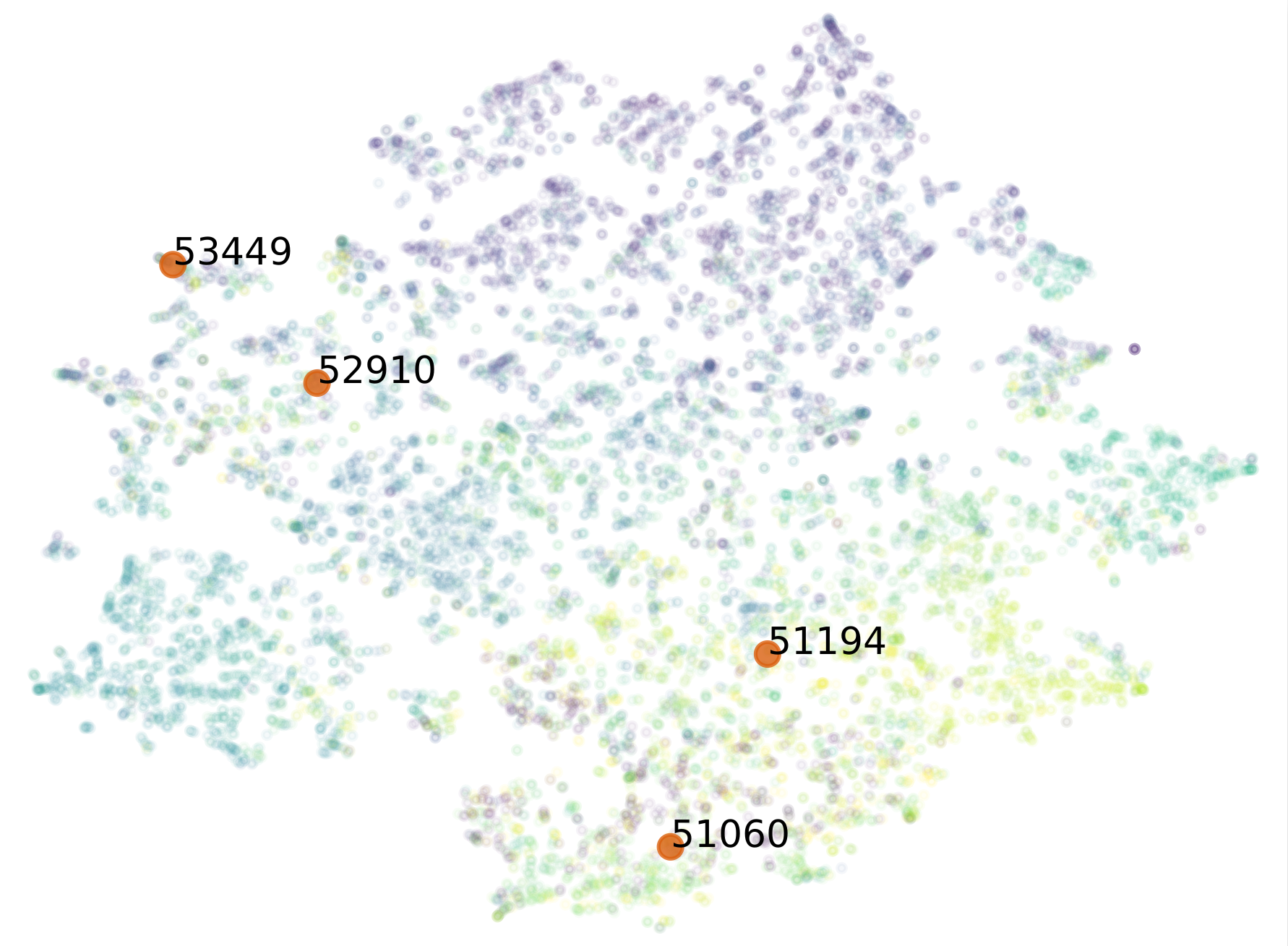}}%
    \caption{The 20 Newsgroups dataset was reduced with t-SNE and LDA as proposed by Atzberger and Cech et al. \cite{benchmarking}. The documents \numprint{51060}, \numprint{51194}, and \numprint{52910}, and \numprint{53449} are highlighted. They are wide-spread in the visualization and are also semantically dissimilar.}
    \label{fig:lda}
\end{figure}

\begin{figure}[t]
    \centering
\begin{lstlisting}[language={},style=newsgroupstyle,linerange={23-53}]
                                   Overview

Welcome to alt.atheism and alt.atheism.moderated.

This is the first in a series of regular postings aimed at new readers of the
newsgroups.

Many groups of a 'controversial' nature have noticed that new readers often
come up with the same questions, mis-statements or misconceptions and post
them to the net.  In addition, people often request information which has
been posted time and time again.  In order to try and cut down on this, the
alt.atheism groups have a series of five regular postings under the following
titles:

   1.  Alt.Atheism FAQ: Overview for New Readers
   2.  Alt.Atheism FAQ: Introduction to Atheism
   3.  Alt.Atheism FAQ: Frequently Asked Questions (FAQ)
   4.  Alt.Atheism FAQ: Constructing a Logical Argument
   5.  Alt.Atheism FAQ: Atheist Resources

This is article number 1.  Please read numbers 2 and 3 before posting.  The
others are entirely optional.

If you are new to Usenet, you may also find it helpful to read the newsgroup
news.announce.newusers.  The articles titled "A Primer on How to Work With
the Usenet Community", "Answers to Frequently Asked Questions about Usenet"
and "Hints on writing style for Usenet" are particularly relevant.  Questions
concerning how news works are best asked in news.newusers.questions.

If you are unable to find any of the articles listed above, see the "Finding
Stuff" section below.
\end{lstlisting}%
    \caption{Document \numprint{52910} has the maximum distance to the supporting document \numprint{51060} after executing t-SNE and LSI as proposed by Atzberger and Cech et al. \cite{benchmarking}.}
    \label{fig:lsi_max}
\end{figure}

\begin{figure}[t]
    \centering
\begin{lstlisting}[language={},style=newsgroupstyle,linerange={11-32}]
In article <1993Apr19.151120.14068@abo.fi> MANDTBACKA@FINABO.ABO.FI (Mats Andtbacka) writes:
>In <930419.125145.9O3.rusnews.w165w@mantis.co.uk> mathew writes:
>> I wonder if Noam Chomsky is reading this?
>
>      I could be wrong, but is he actually talking about outright
>_government_ control of the media, aka censorship?
>
>      If he doesn't, any quick one-stop-shopping reference to his works
>that'll tell me, in short, what he _does_ argue for?

"Manufacturing Consent," a film about the media. You alternative movie source
may have this; or to book it in your local alternative theatre, contact:

FILMS TRANSIT * INTERNATIONAL SALES
Jan Rofekamp
402 Notre Dame E.
Montreal, Quebec
Canada H2Y 1C8
Tel (514) 844-3358 * Fax (514) 844-7298
Telex 5560074 Filmtransmtl

(US readers: call Zeitgeist Films at 212 274 1989.)
\end{lstlisting}%
    \caption{Document \numprint{53449} has the maximum distance to the supporting document \numprint{51060} after executing t-SNE and LDA as proposed by Atzberger and Cech et al. \cite{benchmarking}.}
    \label{fig:lda_max}
\end{figure}

\section{Dataset Interrogation Exemplified}
\label{sec:case study}
In the following we, exemplify the interrogation of datasets using our suggested approach of combining Grounded Theory with Hypothesis Testing through Visualization on the 20 Newsgroups dataset and the MNIST dataset.

\subsection{The 20 Newsgroups dataset}
Studies such as Asim et al.~\cite{asim_et_al}, Chandra~\cite{almost_no_discussion_1}, and Chen and Dai~\cite{CHEN2021253} use the 20 Newsgroups dataset for text classification tasks, consequently implicitly assuming that it constitutes an applicable representative of text datasets.
Thus, we would assume that documents of the same label belong to the same category, which should be reflected by such texts being already similar on the surface.

\paragraph{Grounded Theory.}
To test the assumption, we apply Close Reading to some samples with the lexicographically first class label of the 20 Newsgroups, namely \textit{alt.atheism}.
Derived from the name of the class, we expect that this label represents the category of \textit{atheism}.
For the sake of the argument, let's assume that we randomly collect document \numprint{51060} (shown in \autoref{fig:good_atheism_1}) and find that it contains an introductory article about atheism.
Therefore, it is safe to code this document as being concerned with atheism.
Then, we use Theoretical Sampling to acquire further samples of the label alt.atheism.
Because, our current theory is concerned with all samples of this label, in this specific case we can use an arbitrary sampling method.
We proceed with Coding the next lexicographically ordered 100 samples, during which we obtain document \numprint{51194} (shown in \autoref{fig:good_atheism_2}). 
We find that it is hardly connected to the category of atheism and, therefore, can not be coded as belonging to this category.
Now, we have two options: (1) Based on the current theory that class labels and categories are linked, we could exclude this sample from the texts with the label alt.atheism.
This would require the introduction of a new label for this sample.
(2) Otherwise, we have the option to refine the theory (that documents of the same label belong to the same category) to maintain the same label for both samples.
Given that no reviewed work introduces new categories or codes (see \autoref{tab:review}), we opt for refining the theory and investigate how document \numprint{51060} and document \numprint{51194} can possibly belong to the same category.
By manually reverse engineering the corresponding mailing list thread, we find that document \numprint{51194} is part of a discussion about \textit{if and if true how the growing number of people identifying as atheists is correlated with the growing number of depression cases}. This broader context explains the label of the document. However, as this context, which is not included in the dataset, is necessary for understanding how the document is connected to the atheism discourse in the 80s, the out-of-context document can not be considered as belonging to the category of atheism.
Thus, an ML model trained on the dataset would not learn anything about atheism as understood in 2015 or later, which is the time covered by our previous literature review.

\begin{figure*}[t]
    \centering
    \includegraphics[width=0.085\linewidth]{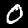}
    \hspace{2pt}
    \includegraphics[width=0.085\linewidth]{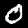}
    \hspace{2pt}
    \includegraphics[width=0.085\linewidth]{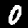}
    \hspace{2pt}
    \includegraphics[width=0.085\linewidth]{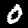}
    \hspace{2pt}
    \includegraphics[width=0.085\linewidth]{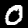}
    \hspace{2pt}
    \includegraphics[width=0.085\linewidth]{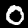}
    \hspace{2pt}
    \includegraphics[width=0.085\linewidth]{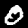}
    \hspace{2pt}
    \includegraphics[width=0.085\linewidth]{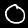}
    \hspace{2pt}
    \includegraphics[width=0.085\linewidth]{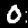}
    \hspace{2pt}
    \includegraphics[width=0.085\linewidth]{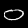}
    \caption{Some examples from the lexicographically first label representing the \emph{0} category from the MNIST dataset. All investigated samples could be coded by us as representing zero even if some artifacts are present, as can be seen in the last example.}
    \label{fig:zeros_mnist}
\end{figure*}

\begin{figure}[t]
    \centering%
    \frame{\includegraphics[width=1.0\columnwidth]{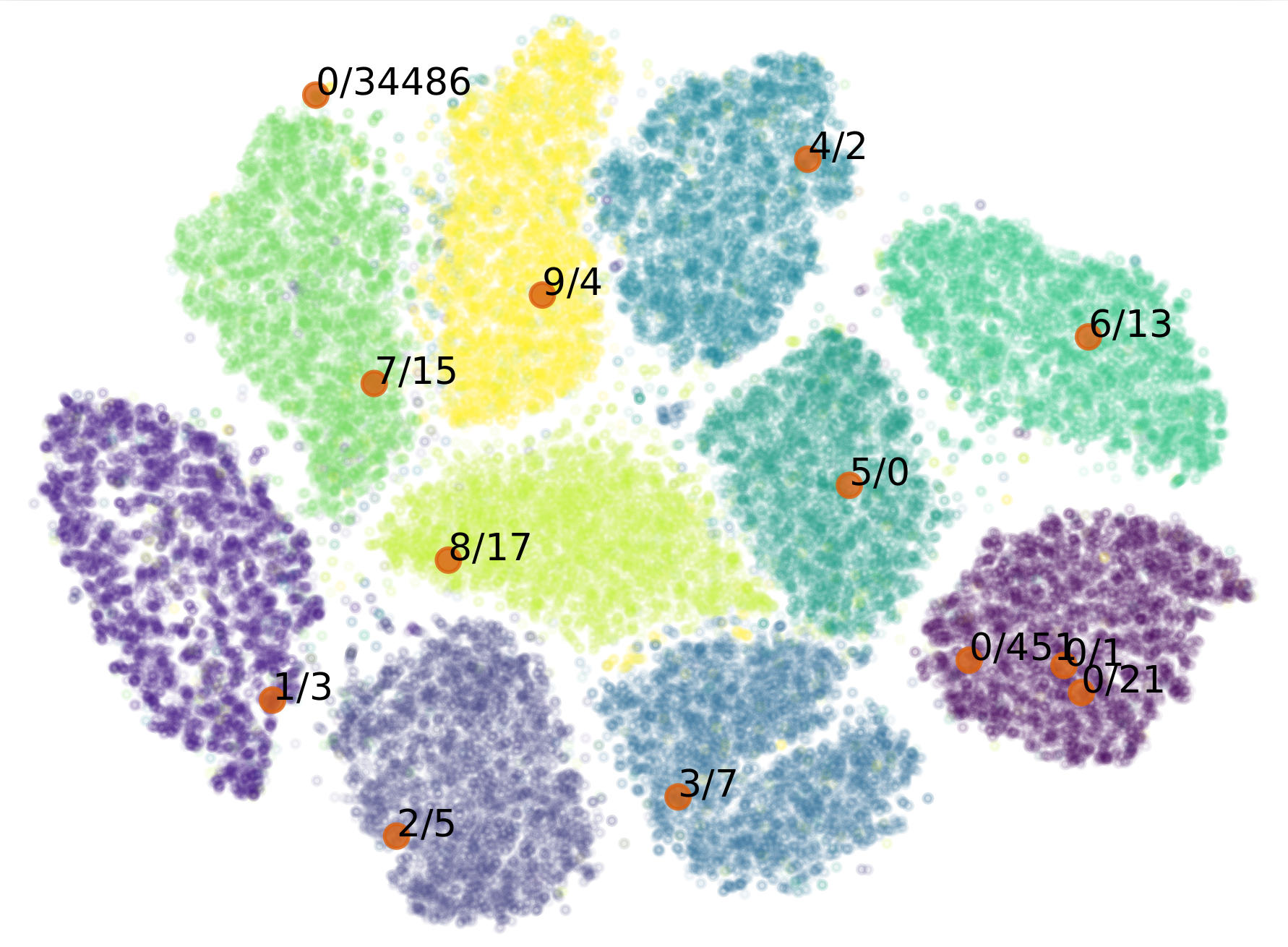}}%
    \caption{An example layout generated according to the guidelines of \cite{espadoto2019toward}. The MNIST dataset was reduced with t-SNE. The images 1, 21, \numprint{451}, and \numprint{34486} of category \emph{0} are highlighted. Even the noisy image 451 clearly belongs to a visually well defined cluster and is semantically similar.}%
    \label{fig:mnist}%
\end{figure}

Further Coding, then, strengthens the theory that in 2015 the documents do not constitute a coherent category any more.
After manually Coding the lexicographically first 100 documents in 11 codes from which only one represents the category of atheism\footnote{\href{https://zenodo.org/record/8337723}{Dataset available with \textsc{doi}: 10.5281/zenodo.8337723}}, we find that only 26 of the documents fit the category of atheism.
Overall, from today's point of view, the only connection between the documents appears to be that they were posted to the same mailing list and not that they belong to the same category with respect to the text content.
Therefore, any model that is trained with the implicit assumption that the label matches the category consequently produces meaningless results and it has to be doubted that the performance of an ML model for text classification on this \enquote{standard dataset} is a good indicator for its general quality.
On the positive side, we found out that the used language in the documents is very diverse and heterogeneous with some documents using informal, short, and heated language and others using very formal and long-winded language which can be considered as real noise due to its independence from the topic of atheism.


\paragraph{Hypothesis Testing through Visualization.}

Having obtained indicators from Grounded Theory that the 20 Newsgroups dataset is actually not suited for learning concepts from textual data, we now choose the opposite of this observation as null hypothesis for the next step.
So we now want to find out, whether we should reject the theory that \textit{the 20 Newsgroups dataset is a good representation for learning concepts from textual data}, by employing Hypothesis Testing through Visualization.
As advised by Atzberger and Cech et al. \cite{benchmarking}, we use \textit{Latent Semantic Indexing (LSI)}~\cite{lsi} and \textit{Latent Dirichlet Allocation (LDA)}~\cite{blei2003latent} for topic modeling and \textit{t-stochastic neighbor embedding (t-SNE)}~\cite{van2008visualizing} for DR.
We choose document \numprint{51060} again as a first representative that supports the proposition, as determined by the previous interrogation.
Subsequently, we apply the guidelines from Atzberger and Cech et al. \cite{benchmarking}, resulting in a two-dimensional representation suitable for investigating the neighborhood of the currently considered supporting document \numprint{51060}.
The resulting scatter plots are rather convoluted (see \autoref{fig:lsi} and \autoref{fig:lda}), as we would expect from the results of the previous section, namely that texts with the same label are actually only loosely connected with respect to their category, if connected at all.
We then observe that the points representing samples from the label alt.atheism are widespread, again suggesting a loosely defined category.
Investigating the immediate neighborhood of document \numprint{51060}, we find document \numprint{51122} as the nearest document using the LDA model and document \numprint{52499} for the respective LSI model.
A manual investigation reveals that those documents are indeed quite well related to the category and therefore strengthen the proposition.
However, document \numprint{51194}, which also previously countered the hypothesis, is not in the neighborhood of document \numprint{51060}, but over \numprint{200} times farther away than the respective closest document mentioned above (according to euclidian distance when using LSI).
On investigation of the documents with the label alt.atheism that are the farthest away from document \numprint{51060} (document \numprint{52910} for LSI and document \numprint{53449} for LDA), we find that their contents are indeed dissimilar from the considered document, as shown in \autoref{fig:lsi_max} and \autoref{fig:lda_max}.
Document \numprint{52910} is an FAQ for the mailing list and is not concerned with atheism and document \numprint{53449} is a comment on a book by Noam Chomsky. 
However, the comment itself has no link to the contemporary category of atheism.
By using multiple different topic models, as proposed by Atzberger and Cech et al. \cite{benchmarking}, we find different types of dissimilarities while still observing that semantically dissimilar documents are dissimilar in both layouts.
The investigated documents counter the \enquote{null hypothesis} and, therefore, we have evidence that the convoluted layout indeed corresponds to imprecise class labels.
Thus, we reject the hypothesis that the labels match the categories on the basis of reasoning over all samples due to the employment of a quantitative DR technique.

\subsection{The MNIST dataset}
We apply the same method as before, starting again with the proposition that the MNIST dataset is a good representation of a computer vision dataset, suitable for benchmark studies on classification of image data, as done by Chen et al. \cite{chen2018assessing}, Garg et al. \cite{garg2019validation}, and Karayaneva and Hintea \cite{karayaneva2018object}.

\paragraph{Grounded Theory.}
We again investigate the lexicographically first label, representing the category \emph{0} by name.
Consequently, we would expect that all images represent zeros, meaning that the category would match the label.
We code the first 100 lexicographically ordered samples with label \emph{0}, by deciding for each image, whether it actually shows a zero or not (see examples in \autoref{fig:zeros_mnist}).
For all cases we can identify the image content as zero, with only rarely encountering minor artifacts, e.g., in the second to last example image in \autoref{fig:zeros_mnist} (image \numprint{451}) or the last zero (image \numprint{34486}), which looks unusual but is still recognizable as a zero.
Since all investigated samples can be coded to the proposed category, we can maintain the proposition that the (class) label \emph{0} matches the category \emph{0}.

\paragraph{Hypothesis Testing through Visualization.}
Espadoto et al.~\cite{espadoto2019toward} identify four DR techniques that work reasonably well for several types of data, among them also t-SNE.
We choose this technique to simplify the comparison to the findings for the 20 Newsgroups dataset.
Starting from the proposition that the MNIST dataset is not a suitable dataset for image classification due to semantically incoherent categories (opposite to the result of applying Grounded Theory), we visualize the dataset for further investigation (see \autoref{fig:mnist}).
In contrast to the 20 Newsgroups dataset, the cluster with label \emph{0} is visually coherent.
Even noisy images, such as image \numprint{451} (the second to last image in \autoref{fig:zeros_mnist}), are located in the dense cluster of zeros.
This observation matches the previous Coding from the Grounded Theory approach well and counters the \enquote{null hypothesis}.
We then investigate one of the few samples that are located in another neighborhood, compared to the rest of the images with the label \emph{0}: image \numprint{34486} (last image in \autoref{fig:zeros_mnist}).
This data point is located in the neighborhood of the visual cluster that represent images containing a seven rather than a zero.
By manual investigation, we find that this data point is indeed dissimilar to the other zeros we manually reviewed but still can be coded as belonging to the category \emph{0}.
Therefore, this case represents \emph{real} noise that is qualitatively similar to the category but quantifiably dissimilar in the underlying value distribution.
Combined with the previous qualitative interrogation, we now have evidence that the visual coherent clusters are not qualitatively distorted by the DR technique.
Therefore, we reject the null hypothesis and strengthen the previous theory that in the case of the MNIST dataset, the labels match the categories.





\subsection{Summary}
By applying the proposed method, we have shown that the alt.atheism label of the 20 Newsgroups dataset is imprecise. Therefore, the data associated with this label is not suitable to learn from, especially in a meaningful fashion.
A model that reaches high accuracy on correctly predicting the alt.atheism class label for a text has not learned a reliable understanding of the category of atheism or any other coherent category.
On the other hand, the MNIST dataset poses no such difficulties.
We found that it has some quantifiable diversity in the class with the label \emph{0}, however, a qualitative review revealed that this noise could still be subsumed under the category \emph{0}.
Therefore, we can conclude that this dataset is a meaningful standard dataset for the intended task of image recognition.
Consequently, and only now, we can further conclude that performance increases from models trained on the MNIST dataset will mean an actually better solution of the use case and that the usual optimization towards accuracy is meaningful.

\section{Discussion}
\label{sec:Discussion}
The investigation of the two exemplary datasets showed how the standardness of a dataset can cloud its actual quality, i.e., whether the labels match the (implicitly) assumed categories. 
Further, data scientists may be misled to think that this standardness implies a fit to their use case, which, however, might have distinct requirements for categories and labels, not satisfied by the dataset at hand.
Therefore, it is not advisable to simply rely on the standardness of a dataset, which typically only refers to its frequent use.
Omitting the discussion about why a dataset is used is only acceptable if the dataset is proven to have a trivial match between labels and categories.
In this case, the concept of \emph{standardness} becomes associated with qualitative criteria, deepening the meaning of comparisons over such a standard dataset.

As shown, the suggested approach enables the assessment of datasets regarding use-case-specific quality criteria. Nonetheless, the approach is subject to threats to validity.
The application of Grounded Theory is time consuming since it requires manual evaluation. 
Furthermore, it requires a clear understanding of the concepts to investigate, otherwise suboptimal codes could be chosen.
In this paper, we were able to validate the concepts without domain experts.
If the use case is more complex and domain-specific, we encourage the involvement of domain experts to mitigate the risk of suboptimal Coding.
A further limitation is that for every class label a lot of samples would have to be investigated until Theoretical Saturation is achieved.
This is usually not feasible for a dataset with at least several thousands of samples.
This is the reason why, in addition to Grounded Theory, we proposed a more quantitative method with Hypothesis Testing through Visualization to help overcoming this limitation by analyzing many samples at once.
Unfortunately, every DR for a non-trivial dataset leads to some sort of distortion of the high-dimensional structure~\cite{jeon2023classes}.
Therefore the observed patterns in the two-dimensional plane may not correspond to patterns in the original high-dimensional space.
We mitigate this risk by using the result of large benchmark studies~\cite{benchmarking, espadoto2019toward} and combining it with the results from the Grounded Theory approach.
This way, the qualitative approach mitigates some of the risks of the quantitative approach and vice versa~\cite{quali_quantitative_methods}.
Using a quali-quantitative approach, enables to reveal evidence which would remain hidden by a purely quantitative approach or not reasonably obtainable in time by a purely qualitative approach.


\section{Conclusions}
\label{sec:Conclusions}
The importance of dataset quality and suitability is nowadays well known in the field of ML. 
Nevertheless, we observed that for standard datasets these issues are rarely taken into account. 
If the match between use case, derived categories, and class labels had been connected with the standardness of a dataset, trustworthy and meaningful results could be obtained from an ML model trained on this dataset.
However, the degree of this match is not guaranteed and often not discussed, as we demonstrated by reviewing recent literature.
Here we saw that researchers often seem to equate standardness with quality and suitability, even though standardness is often solely a result of historical developments and usage frequency.
Therefore, we encourage researchers to be more critical with their use of standard datasets and assess the quality and suitability of the dataset.
For this, we suggested a quali-quantiative approach to dataset interrogation and demonstrated its application in detail.

As a promising research direction, we perceive the (interactive) refinement of existing datasets, e.g., using approaches as the one from Chatzimparmpas et al.~\cite{HardVis}.
Data scientists may want to introduce new labels to their datasets, e.g., by the use of label propagation approaches in low dimensional spaces~\cite{Benato_1}, approaches to incorporate domain knowledge into a DR~\cite{human_informed_dr}, or reverse engineering and correction of assumptions that were prevalent during gathering of a (standard) dataset~\cite{chhavi_reconstruction}.
Further, our suggested method for manual dataset assessment would benefit from methods for determining when a DR can be reliably used, beyond mere quantitative benchmark studies.
For example, one could imagine an analogon to the p-value for two-dimensional DR layouts as another step towards the combination of human reasoning with the abstraction potential provided by artificial intelligence and visualization methods.

\bibliographystyle{abbrv-doi-hyperref-narrow}

\bibliography{main}


\end{document}